\documentclass[letterpaper]{article}
\usepackage[preprint]{aaai2027}
\usepackage[hyphens]{url}
\usepackage{graphicx}
\usepackage{natbib}
\usepackage{caption}
\usepackage{algorithm}
\usepackage{algorithmic}
\usepackage{amsmath}
\usepackage{multirow}
\usepackage[table]{xcolor}
\usepackage{amsfonts}
\usepackage{amsthm}

\newtheorem{theorem}{Theorem}
\newtheorem{remark}{Remark}
\newtheorem{lemma}{Lemma}
\usepackage{newfloat}
\usepackage{listings}
\DeclareCaptionStyle{ruled}{labelfont=normalfont,labelsep=colon,strut=off} 
\floatstyle{ruled}
\newfloat{listing}{tb}{lst}{}
\floatname{listing}{Listing}
\usepackage{booktabs}
\title{TEMPER: Tensorized Efficient Manifold-constrained Parameterization for Expressive Residual Routing}
\author{
    Yuxuan Gu\textsuperscript{\rm 1}, 
    Wuyang Zhou \textsuperscript{\rm 1},
    Huijun Xing \textsuperscript{\rm 1},
    Danilo Mandic \textsuperscript{\rm 1}
}

\affiliations{
    \textsuperscript{\rm 1}Department of Electrical and Electronic Engineering, Imperial College London\\ \{yuxuan.gu21, wuyang.zhou19, h.xing23@imperial.ac.uk, d.mandic\}@imperial.ac.uk
}

\begin{document}

\maketitle

\begin{abstract}
Residual connections rely on a static residual pathway, and are essential for training deep neural networks. Hyper-connections (HC) increase the expressivity of residual routing  by incorporating multiple residual streams and learning dynamic information flow, while manifold-constrained (mHC) variants stabilize training through doubly stochastic residual mixing. However, a generator-level bottleneck remains in existing methods: they use dense, unstructured generators for pre-branch aggregation, residual mixing, and post-branch redistribution, which results in parameter count growing rapidly with the number of streams. To address this issue, we propose \underline{\textbf{T}}ensorized 
\underline{\textbf{E}}fficient \underline{\textbf{M}}anifold-constrained \underline{\textbf{P}}arameterization for \underline{\textbf{E}}xpressive Residual \underline{\textbf{R}}outing (\textbf{TEMPER}), which represents these generators as multi-way tensors over the input-stream, feature, and output-stream modes, and parameterizes them using tensor networks. Such a structured low-rank formulation is shown to preserve token-dependent manifold-constrained routing interface while substantially reducing parameter growth. It also promotes interpretability and intuition, as: i) tensor ranks control the dimensionality of the learned routing subspace, with full ranks recovering dense routing; while ii) the generator approximation errors bound differences in routing logits and, consequently, in the routed-block outputs. Comprehensive experiments show that TEMPER matches or outperforms existing methods across language modeling and commonsense reasoning tasks, while requiring substantially fewer additional parameters. At eight residual streams, TEMPER achieves the best CORE score while using about $84\%$ fewer additional parameters than mHC, thus showing a stronger performance-parameter efficiency trade-off.

\end{abstract}

\section{Introduction}

Residual connections \cite{he2016deep} are central to the stability and scalability of modern Transformers \cite{vaswani2017attention,NEURIPS2020_1457c0d6}. In a standard residual block, a module computes an update which is added back to the token's current representation. This design is simple and effective, but it routes every token via the same fixed residual path.

\begin{figure}[ht]
    \centering
    \includegraphics[width=\columnwidth]{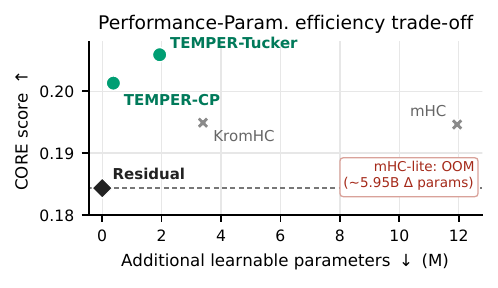}
    \caption{Comparison of CORE score and additional parameter cost at $n=8$. Our TEMPER method provides the strongest trade-off among all hyper-connection variants.}
    \label{fig:pareto_teaser}
\end{figure}

Recently, hyper-connections (HC) \cite{zhu2025hyper} have enriched this design by maintaining multiple residual streams and learning token-dependent routing among them, thus enabling more complex residual topologies and information flow patterns. Instead of applying an identical and fixed residual connection to all tokens, these methods allow each token to adaptively aggregate, mix, and redistribute information across several streams. Manifold-constrained hyper-connections (mHC) \cite{xie2026mhc} further improve the training stability of HC by using the iterative Sinkhorn-Knopp (SK) optimization \cite{sinkhorn1964relationship} to constrain the residual mixing matrix, $\mathbf{A}^{\mathrm{res}}$, to be doubly stochastic. Subsequent variants have reparameterized this residual mixing step. Among these, the mHC-lite method \cite{yang2026mhc} achieves exact doubly stochastic residual mixing by constructing $\mathbf{A}^{\mathrm{res}}$ as a convex combination of permutation matrices \cite{birkhoff-von-neumann}, while KromHC \cite{zhou2026kromhc} uses Kronecker products of smaller doubly stochastic matrices \cite{taranenko2023products}, in order to avoid factorial growth of parameters with the number of streams incurred by mHC-lite. Figure~\ref{fig:pareto_teaser} summarizes the performance-parameter trade-offs of the existing methods and the proposed TEMPER, and highlights the remaining parameter efficiency gap.

Although these HC variants improve the parameterization of the residual mixing matrix, the generator-level bottleneck still remains an open issue. For example, mHC uses unstructured dense generators to produce all three token-dependent routing projectors from stream-feature representations. As the number of streams increases, these generators become increasingly expensive in terms of both parameter count and computation, limiting the scalability of the expressive token-dependent routing. This raises the following question:

\begin{figure*}[ht]
\centering
\includegraphics[width=\linewidth]{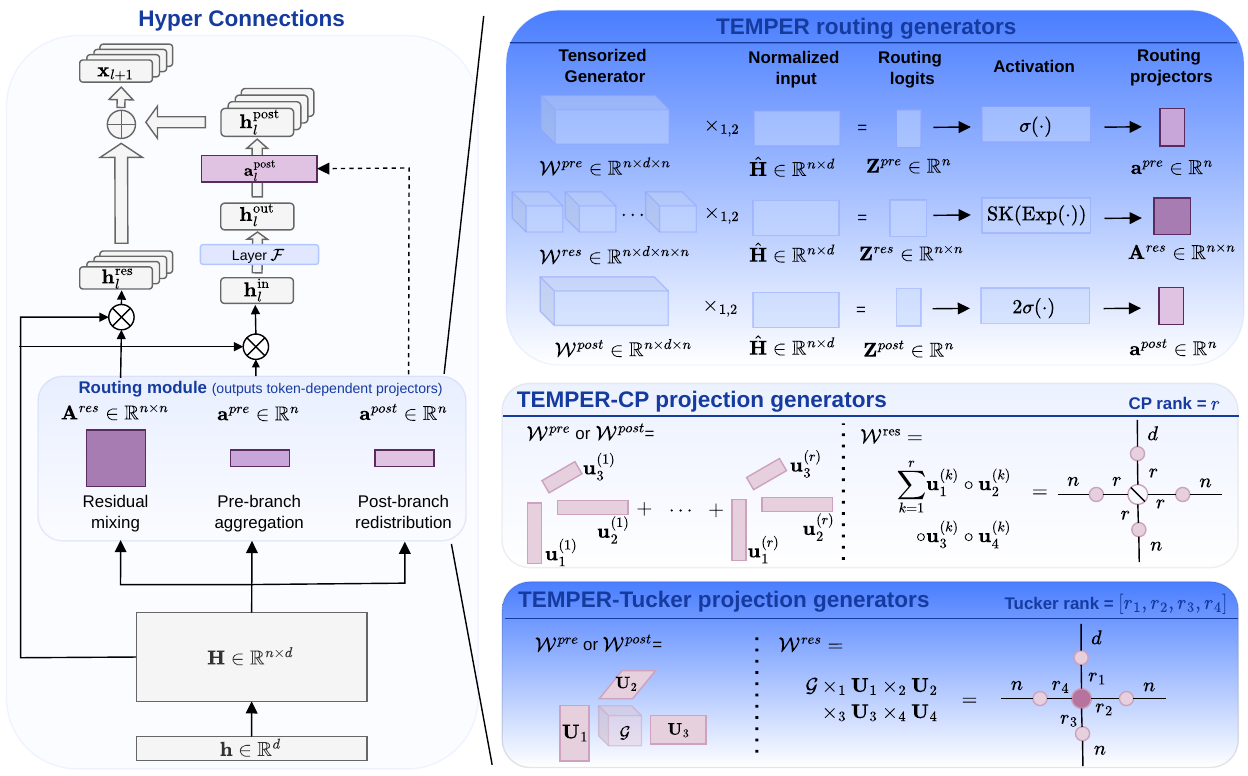}
\caption{Overview of TEMPER architecture. TEMPER replaces unstructured dense hyper-connections routing generators with tensorized TEMPER-CP or TEMPER-Tucker generators, while preserving the same token-dependent routing interfaces.}
\label{fig:architecture}
\end{figure*}

\begin{itemize}
    \item \textit{Can we replace the dense unstructured routing generators in mHC with structured, parameter-efficient alternatives, while achieving on par or even better performance?}
\end{itemize}

To answer this question, we propose \underline{\textbf{T}}ensorized 
\underline{\textbf{E}}fficient \underline{\textbf{M}}anifold-constrained \underline{\textbf{P}}arameterization for \underline{\textbf{E}}xpressive Residual \underline{\textbf{R}}outing (\textbf{TEMPER}), which replaces the dense routing generators with parameter-efficient tensor-network parameterizations. By exploiting the inherently multi-way structure of mHC routing, TEMPER represents the generators for pre-branch aggregation, residual mixing, and post-branch redistribution as higher-order tensors over their input-stream, feature, and output-stream modes, and factorizes each independently using a low-rank tensor network.

TEMPER retains the original token-dependent routing interface, including doubly stochastic residual mixing, while inducing structured parameter sharing within each generator across stream–feature–output interactions. Its tensor ranks provide an explicit capacity control. Compact ranks restrict routing to low-dimensional learned subspaces, whereas sufficiently large ranks recover the dense parameterization. TEMPER thus reduces routing-generator parameter growth without altering the surrounding hyper-connection architecture.

Empirically, we instantiate TEMPER with Canonical Polyadic (CP) \cite{hitchcock1927expression} and Tucker \cite{tucker1966some} parameterizations. Both outperform prior hyper-connection baselines in downstream performance while using substantially fewer additional parameters. At eight residual streams, TEMPER-Tucker obtains the best CORE score ($0.206$ versus $0.195$ for both mHC and KromHC) with $1.93$M additional parameters, which is $84\%$ fewer than $11.95$M in mHC.

In summary, our contributions are as follows:
\begin{itemize}
    \item We introduce \textbf{TEMPER}, a tensorized manifold-constrained hyper-connections method that replaces dense token-dependent routing generators with structured efficient tensor-network parameterizations, while preserving the original mHC routing interface.

    \item We analyze TEMPER's scaling and expressivity. At fixed
     rank, routing-generator parameter grows linearly with the number of streams. Full-rank TEMPER-Tucker recovers dense routing, while generator approximation errors bound routing-logit and block-output deviations.
    
    \item We instantiate TEMPER with CP and Tucker parameterizations and
    evaluate them on language modeling and commonsense reasoning.
    TEMPER achieves a stronger performance--parameter efficiency trade-off
    than the existing residual and hyper-connection baselines.
\end{itemize}

\section{Related Work}\label{sec:related_work}

In a Transformer block $f(\cdot)$, standard residual connections \cite{he2016deep} update every token's latent representation using the same pre-defined residual pathway
\begin{equation}
    \mathbf{h}' = \mathbf{h} + f(\mathbf{h}),
\end{equation}
where $\mathbf{h}$ is the hidden representation. While this design is computationally efficient and stable in optimization, it imposes the identical information flow pattern across all tokens. 

\subsection{Hyper-Connections Variants}

\paragraph{Hyper-connections (HC)} \cite{zhu2025hyper} address this limitation by expanding the residual path to $n$ streams and introducing token-specific routing over these streams. This increases the topological and representational complexity of the residual path,
enabling richer and more diverse feature interactions across streams.

Specifically, at a block $l$, each token's representation is repeated $n$ times to obtain the expanded state matrix \footnote{The superscript $(l)$ is omitted for clarity from here onwards.}
$
    \mathbf{H}^{(l)} =
    \begin{bmatrix}
        \mathbf{h}_1^{(l)\top}, \ldots, \mathbf{h}_n^{(l)\top}\\
    \end{bmatrix}^{\top} \in \mathbb{R}^{n \times d}
$. A routing module produces three token-dependent projection matrices
$\mathbf{a}^{\mathrm{pre}} \in \mathbb{R}^{1 \times n}$,
$\mathbf{A}^{\mathrm{res}} \in \mathbb{R}^{n \times n}$, and
$\mathbf{a}^{\mathrm{post}} \in \mathbb{R}^{1 \times n}$, with the resulting update
\begin{equation}
    \mathbf{H}'
    =
    \mathbf{A}^{\mathrm{res}}\mathbf{H}
    +
    (\mathbf{a}^{\mathrm{post}})^T
    f\!\left(\mathbf{a}^{\mathrm{pre}}\mathbf{H}\right).
    \label{eq:stream-update}
\end{equation}
Thus, the residual mixing is no longer globally fixed. Instead, it is selected dynamically for each token.

\paragraph{Manifold-Constrained Hyper-Connections (mHC)} \cite{xie2026mhc} constrains the residual mixing matrix, $\mathbf{A}^{\mathrm{res}}$, to be doubly stochastic, since an unconstrained matrix can disrupt the identity-mapping behavior which is the key to training stability in standard residual networks. Let
$\hat{\mathbf{h}}=\mathrm{vec}(\mathrm{RMSNorm}(\mathbf{H}))\in \mathbb{R}^{1 \times nd}$
be the normalized flattened state. Then, mHC generates the routing parameters through
\begin{equation}
    \begin{aligned}
        \mathbf{a}^{\mathrm{pre}}
        &=
        \sigma\!\left(
        \alpha^{\mathrm{pre}}\hat{\mathbf{h}}\mathbf{W}^{\mathrm{pre}}
        +
        \mathbf{b}^{\mathrm{pre}}
        \right), \\
        \mathbf{A}^{\mathrm{res}}
        &=
        \mathrm{SK}\!\left(
        \exp\!\left(
        \mathrm{mat}\!\left(
        \alpha^{\mathrm{res}}\hat{\mathbf{h}}\mathbf{W}^{\mathrm{res}}
        +
        \mathbf{b}^{\mathrm{res}}
        \right)\right)\right). \\
        \mathbf{a}^{\mathrm{post}}
        &=
        2\sigma\!\left(
        \alpha^{\mathrm{post}}\hat{\mathbf{h}}\mathbf{W}^{\mathrm{post}}
        +
        \mathbf{b}^{\mathrm{post}}
        \right), \\
    \end{aligned}
\end{equation}
Here,
$
\mathbf{W}^{\mathrm{pre}}, \mathbf{W}^{\mathrm{post}} \in \mathbb{R}^{nd \times n},
\mathbf{b}^{\mathrm{pre}}, \mathbf{b}^{\mathrm{post}} \in \mathbb{R}^{1 \times n},
\mathbf{W}^{\mathrm{res}} \in \mathbb{R}^{nd \times n^{2}},
\mathbf{b}^{\mathrm{res}} \in \mathbb{R}^{1 \times n^{2}},
$ $\alpha^{\mathrm{pre}}, \alpha^{\mathrm{post}}, \alpha^{\mathrm{res}} \in \mathbb{R}$, and $\sigma(\cdot)$ denotes the $\mathrm{Sigmoid}(\cdot)$ function. Moreover, $\mathrm{SK}(\cdot)$ denotes the Sinkhorn--Knopp operator \cite{sinkhorn1964relationship}, which iteratively normalizes $\mathbf{A}^{\mathrm{res}}$ towards a doubly stochastic matrix.
As a result, residual mixing becomes a non-expansive convex combination over streams, which improves training stability.

\paragraph{mHC-lite and KromHC.}
The mHC-lite method \cite{yang2026mhc} replaces iterative SK normalization with a
Birkhoff--von Neumann mixture, while KromHC \cite{zhou2026kromhc} factorizes
that mixture using smaller Kronecker factors. Both leave the pre- and
post-routing generators unchanged. Their residual mixers can be summarized as
\begin{equation}
\begin{aligned}
\mathbf{A}^{\mathrm{res}}_{\mathrm{lite}}
&=\sum_{m=1}^{n!}\pi_m\mathbf{P}_m,\\
\mathbf{A}^{\mathrm{res}}_{\mathrm{Krom}}
&=\bigotimes_{k=K}^{1}\mathbf{U}^k,
\qquad
\mathbf{U}^k=\sum_{m=1}^{i_k!}\pi_m^k\mathbf{P}_m^k,
\end{aligned}
\end{equation}
where $n=\prod_{k=1}^K i_k$, and $\boldsymbol{\pi}$ and
$\boldsymbol{\pi}^k$ are token-dependent SoftMax weights over the
corresponding permutation matrices. Both constructions are exactly doubly
stochastic, but mHC-lite requires $\mathcal{O}(n!)$ mixture components,
whereas KromHC reduces this cost via smaller Kronecker factors.
\subsection{Tensor Methods in Large Language Models (LLMs)}

Tensor decompositions, by their nature, improve parameter efficiency across LLM components. The Tensor Train decomposition \cite{oseledets2011tensor} compresses dense layers \citep{novikov2015tensorizing}, while TensorGPT \citep{xu2023tensorgpt} extends this approach to token embeddings. TensorLLM \citep{gu2025tensorllm} and Tucker Attention \citep{klein2026tucker} apply Tucker-style factorizations to multi-head attention. In parameter-efficient fine-tuning, LoRETTA \citep{yang-etal-2024-loretta} uses Tensor Train adapters, whereas TeRA \citep{gu2026tera} enables high-rank adaptation with a Tucker-like network.

The performance of tensorized models depends critically on their ranks and network structures. \citet{zhou2025understanding} study tensor-network ranks as design parameters, while \citet{zeng2024tngps} and \citet{iacovides2025domain} explore domain-aware tensor-network structure search with LLM assistance.

\section{Preliminaries}\label{sec:preliminaries}

\paragraph{Notation.}
Scalars, vectors, matrices, and tensors are denoted by $x$, $\mathbf{x}$, $\mathbf{X}$, and $\mathcal{X}$, respectively \cite{kolda_tensor, cichocki2015tensor, cichocki2016tensor}. We use $(\cdot)^\top$ for matrix transpose, $\mathcal{A}(i_1,\ldots,i_N)$ for the $(i_1,\ldots,i_N)$-th tensor entry, $\times_n$ for the mode-$n$ product, $\circ$ for the outer product, and $\otimes$ for the Kronecker product. The operators $\operatorname{vec}(\cdot)$ and $\operatorname{mat}(\cdot)$ denote the vectorization and matricization, respectively.

\paragraph{Tensor Contraction.}

Contractions are used to sum over shared modes of compatible tensors. For
$\mathcal{A}\in\mathbb{R}^{I_1\times I_2\times\cdots\times I_N}$ and
$\mathbf{X}\in\mathbb{R}^{I_1\times I_2}$, the contraction over the first two modes is
\begin{equation}
    (\mathcal{A}\times_{1,2}\mathbf{X})_{i_3,\ldots,i_N}
    =
    \sum_{u=1}^{I_1}\sum_{v=1}^{I_2}
    \mathcal{A}_{u,v,i_3,\ldots,i_N}\mathbf{X}_{u,v}.
\end{equation}
For $\mathcal{A}$ and $\mathcal{B}$ with mode $a$ of $\mathcal{A}$ and mode $b$ of $\mathcal{B}$ both of size $R$, tensor--tensor contraction
\begin{equation}
    (\mathcal{A}\times_{a}^{b}\mathcal{B})_{i_1,\ldots,i_{N-1},\,j_1,\ldots,j_{M-1}}
    =
    \sum_{t=1}^{R}
    \mathcal{A}_{i_1,\ldots,i_{N-1},t}\mathcal{B}_{t,j_1,\ldots,j_{M-1}},
\end{equation}
with all other modes remaining uncontracted.

\paragraph{Unfolding.}

Unfolding, or matricization, reshapes a tensor into a matrix by mapping the indices of selected modes to the matrix rows and the indices of the remaining modes to the matrix columns. For
$\mathcal{A}\in\mathbb{R}^{I_1\times I_2\times\cdots\times I_N}$, the unfolding that groups the first two modes against the remaining modes is given by
\begin{equation}
    \mathrm{mat}(\mathcal{A})
    \in
    \mathbb{R}^{I_1 I_2\times \prod_{m=3}^{N} I_m},
\end{equation}
while the entries of the unfolded matrix are given by
\begin{equation}
    \mathrm{mat}(\mathcal{A})_{
    (i_1-1)I_2+i_2,\;
    1+\sum_{m=3}^{N}(i_m-1)\prod_{\ell=m+1}^{N} I_{\ell}}
    =
    \mathcal{A}_{i_1,i_2,\ldots,i_N}.
\end{equation}

\section{Methodology}

Existing hyper-connection variants mainly differ in how they constrain or parameterize the residual mixing matrix, $\mathbf{A}^{\mathrm{res}}$. Our focus is on re-designing the dense routing generators through the lens of tensor factorization. The overall architecture is illustrated in Figure \ref{fig:architecture}.

\subsection{A Tensor View of Dense Routing Generators}
We now focus on the generators that produce the routing projectors. While prior methods differ in their construction of $\mathbf{A}^{\mathrm{res}}$, they all produce $\mathbf{a}^{\mathrm{pre}}$, $\mathbf{A}^{\mathrm{res}}$, and $\mathbf{a}^{\mathrm{post}}$ using dense maps. Our TEMPER replaces these dense generators with tensorized maps, while preserving the same routing interface.

Let $\hat{\mathbf{H}}=\mathrm{RMSNorm}(\mathbf{H}) \in \mathbb{R}^{n \times d}$ denote the normalized stream state. Instead of flattening $\hat{\mathbf{H}}$ into a length-$nd$ vector, we retain its stream and feature modes. The dense generators can then be viewed as the unfolding of tensors
$\mathcal{W}^{\mathrm{pre}} \in \mathbb{R}^{n \times d \times n}$, $\mathcal{W}^{\mathrm{post}} \in \mathbb{R}^{n \times d \times n}$, and $\mathcal{W}^{\mathrm{res}} \in \mathbb{R}^{n \times d \times n \times n}$, where the first two modes correspond to the input-streams and features, and the remaining modes index the routing output-streams dimensions. With this tensor view, dense mHC generators can be re-written as
\begin{equation}
    \begin{aligned}
        \mathbf{a}^{\mathrm{pre}}
        &=
        \sigma\!\Bigl(
        \alpha^{\mathrm{pre}}
        \mathcal{Z}^{\mathrm{pre}}
        +
        \mathbf{b}^{\mathrm{pre}}
        \Bigr), \\
        \mathbf{A}^{\mathrm{res}}
        &=
        \mathrm{SK}\!\Bigl(
        \exp\!\Bigl(
        \alpha^{\mathrm{res}}
        \mathcal{Z}^{\mathrm{res}}
        +
        \mathrm{mat}(\mathbf{b}^{\mathrm{res}})
        \Bigr)\Bigr), \\
        \mathbf{a}^{\mathrm{post}}
        &=
        2\sigma\!\Bigl(
        \alpha^{\mathrm{post}}
        \mathcal{Z}^{\mathrm{post}}
        +
        \mathbf{b}^{\mathrm{post}}
        \Bigr),
    \end{aligned}
\label{eq:tensorized_routing}
\end{equation}
where the three routing logits $\mathcal{Z}^{\mathrm{pre}}, \mathcal{Z}^{\mathrm{res}}, \mathcal{Z}^{\mathrm{post}}$ are obtained by contracting the input-stream and feature
modes as
\begin{equation}
\begin{aligned}
    \mathcal{Z}^{\mathrm{pre}}
    &=\mathcal{W}^{\mathrm{pre}}\times_{1,2}\hat{\mathbf{H}},&
    \mathcal{Z}^{\mathrm{pre}}_i
    &=
    \sum_{s=1}^{n}\sum_{c=1}^{d}
    \hat{H}_{s,c}
    \mathcal{W}^{\mathrm{pre}}_{s,c,i},\\
    \mathcal{Z}^{\mathrm{res}}
    &=\mathcal{W}^{\mathrm{res}}\times_{1,2}\hat{\mathbf{H}},&
    \mathcal{Z}^{\mathrm{res}}_{i,j}
    &=
    \sum_{s=1}^{n}\sum_{c=1}^{d}
    \hat{H}_{s,c}
    \mathcal{W}^{\mathrm{res}}_{s,c,i,j},\\
    \mathcal{Z}^{\mathrm{post}}
    &=\mathcal{W}^{\mathrm{post}}\times_{1,2}\hat{\mathbf{H}},&
    \mathcal{Z}^{\mathrm{post}}_i
    &=
    \sum_{s=1}^{n}\sum_{c=1}^{d}
    \hat{H}_{s,c}
    \mathcal{W}^{\mathrm{post}}_{s,c,i}.
\end{aligned}
\label{eq:routing-logits}
\end{equation}

\begin{remark}
    The tensor view in Eq.~\eqref{eq:tensorized_routing} and \eqref{eq:routing-logits} is equivalent to the original dense generators as matricizing each generator tensor recovers its corresponding dense weight
matrix. TEMPER changes only how these tensors are parameterized.
\end{remark}

Thus, dense routing learns independent parameters for each stream-feature-output interaction, with no explicit sharing across modes. This can be inefficient when routing decisions depend on reusable shared stream-feature-output patterns. TEMPER introduces such parameter sharing by factorizing the generator tensors with structured tensor networks.

\subsection{Tensorizing the Routing Generators}
The tensorization is achieved by keeping the contractions in Eq.~\eqref{eq:routing-logits} fixed and by replacing
each dense generator tensor with a structured CP or Tucker tensor factorization.

\paragraph{TEMPER-CP.}
TEMPER-CP uses CP decomposition to factorize each generator tensor into rank-one outer products
\begin{equation}
\begin{aligned}
\mathcal{W}^{\mathrm{pre}}
&=\sum_{q=1}^{r}\mathbf{u}^{\mathrm{pre}}_q\circ\mathbf{c}^{\mathrm{pre}}_q\circ\mathbf{v}^{\mathrm{pre}}_q,\\
\mathcal{W}^{\mathrm{res}}
&=\sum_{q=1}^{r}\mathbf{u}^{\mathrm{res}}_q\circ\mathbf{c}^{\mathrm{res}}_q\circ\mathbf{v}^{\mathrm{res}}_{1,q}\circ\mathbf{v}^{\mathrm{res}}_{2,q},\\
\mathcal{W}^{\mathrm{post}}
&=\sum_{q=1}^{r}\mathbf{u}^{\mathrm{post}}_q\circ\mathbf{c}^{\mathrm{post}}_q\circ\mathbf{v}^{\mathrm{post}}_q.
\end{aligned}
\label{eq:cp-generator}
\end{equation}

Here, $\mathbf{u}_q \in \mathbb{R}^{n}$ and $\mathbf{c}_q \in \mathbb{R}^{d}$ are factors for the input-stream and feature mode, respectively, 
while $\mathbf{v}_q \in \mathbb{R}^{n}$ is the output-stream mode factor. 
For $\mathcal{W}^{\mathrm{res}}$, $\mathbf{v}_{1,q}, \mathbf{v}_{2,q} \in \mathbb{R}^{n}$ denote factors for its two output-stream modes. Within each generator, 
the component index, $q$, couples the mode-specific factors in each rank-one term, inducing structured parameter sharing across stream-feature-output interactions.

\paragraph{TEMPER-Tucker.}
TEMPER-Tucker employs Tucker decomposition to factorize each generator tensor using factor matrices and a small tensor core as
\begin{equation}
\begin{aligned}
\mathcal{W}^{\mathrm{pre}}
&=\mathcal{G}^{\mathrm{pre}}\times_1\mathbf{U}^{\mathrm{pre}}\times_2\mathbf{C}^{\mathrm{pre}}\times_3\mathbf{V}^{\mathrm{pre}},\\
\mathcal{W}^{\mathrm{res}}
&=\mathcal{G}^{\mathrm{res}}\times_1\mathbf{U}^{\mathrm{res}}\times_2\mathbf{C}^{\mathrm{res}}\times_3\mathbf{V}^{\mathrm{res}}_1\times_4\mathbf{V}^{\mathrm{res}}_2,\\
\mathcal{W}^{\mathrm{post}}
&=\mathcal{G}^{\mathrm{post}}\times_1\mathbf{U}^{\mathrm{post}}\times_2\mathbf{C}^{\mathrm{post}}\times_3\mathbf{V}^{\mathrm{post}}.
\end{aligned}
\label{eq:tucker-generator}
\end{equation}
We use Tucker ranks $(r_n,r_d,r_n)$ for the pre- and post-branch generators and $(r_n,r_d,r_n,r_n)$ for the residual generator, with input- and output-stream factors $\mathbf{U}, \mathbf{V} \in \mathbb{R}^{n\times r_n}$ and feature factors $\mathbf{C} \in \mathbb{R}^{d\times r_d}$. Given $\hat{\mathbf{H}}\in\mathbb{R}^{n\times d}$, the input is first projected into a compressed stream--feature representation:
\begin{equation}
\begin{aligned}
\mathbf{T}^{\mathrm{pre}}&=(\mathbf{U}^{\mathrm{pre}})^\top\hat{\mathbf{H}}\mathbf{C}^{\mathrm{pre}},\\
\mathbf{T}^{\mathrm{res}}&=(\mathbf{U}^{\mathrm{res}})^\top\hat{\mathbf{H}}\mathbf{C}^{\mathrm{res}},\\
\mathbf{T}^{\mathrm{post}}&=(\mathbf{U}^{\mathrm{post}})^\top\hat{\mathbf{H}}\mathbf{C}^{\mathrm{post}}.
\end{aligned}
\label{eq:tucker_compression}
\end{equation}
The routing logits are then obtained by contracting this representation with the core and the output modes:
\begin{equation}
\begin{aligned}
\mathcal{Z}^{\mathrm{pre}}
&=(\mathcal{G}^{\mathrm{pre}}\times_{1,2}\mathbf{T}^{\mathrm{pre}})\times_1\mathbf{V}^{\mathrm{pre}},\\
\mathcal{Z}^{\mathrm{res}}
&=(\mathcal{G}^{\mathrm{res}}\times_{1,2}\mathbf{T}^{\mathrm{res}})\times_1\mathbf{V}^{\mathrm{res}}_1\times_2\mathbf{V}^{\mathrm{res}}_2,\\
\mathcal{Z}^{\mathrm{post}}
&=(\mathcal{G}^{\mathrm{post}}\times_{1,2}\mathbf{T}^{\mathrm{post}})\times_1\mathbf{V}^{\mathrm{post}}.
\end{aligned}
\label{eq:tucker_logits}
\end{equation}
TEMPER-Tucker therefore constrains routing to depend on a compressed stream-feature representation rather than on an unrestricted dense map.

\begin{figure}[t]
\centering
\includegraphics[width=\columnwidth]{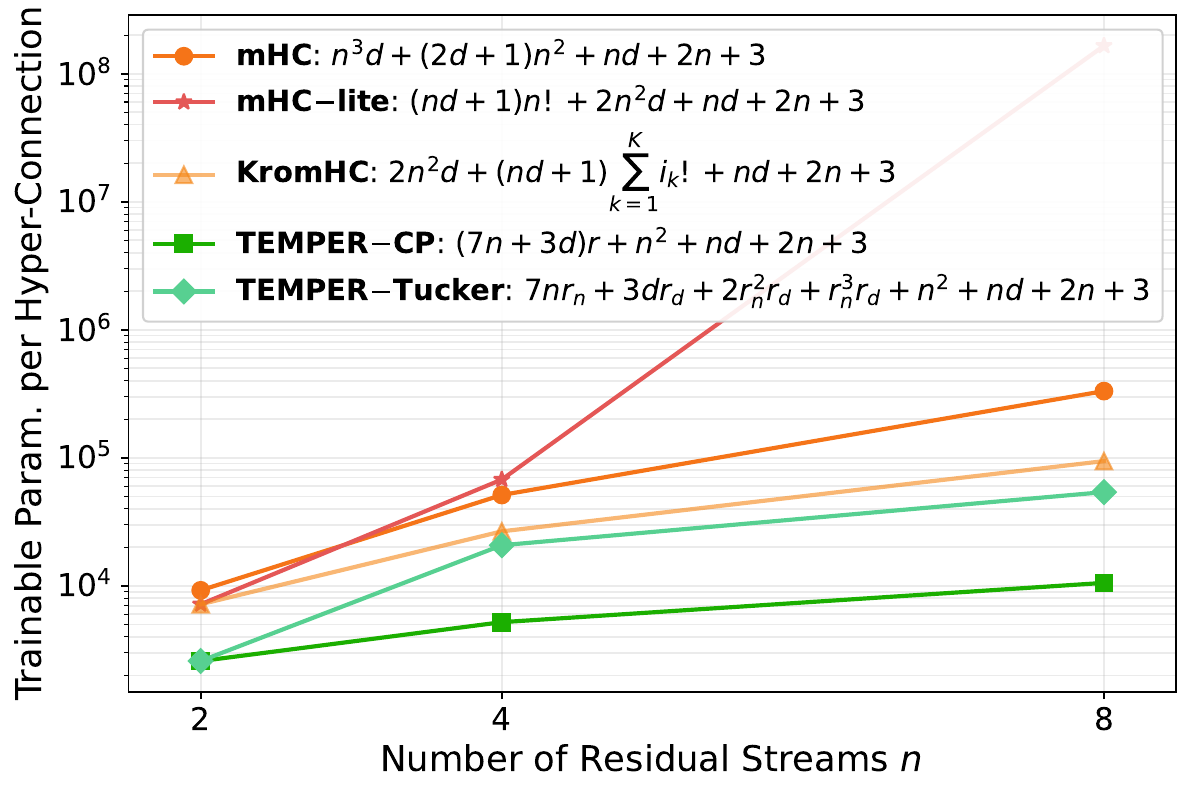}
\caption{Trainable parameters per hyper-connection, including routing generators, biases, scalar gates, and RMSNorm. TEMPER scales more favorably with the number of streams.}
\label{fig:hyperconn_param_efficiency}
\end{figure}

\paragraph{Parameter Efficiency.}
Factorizing the routing generators substantially reduces the parameter cost of
token-dependent routing. The pre-, residual-, and post-routing outputs lie in $\mathbb{R}^n$, $\mathbb{R}^{n\times n}$, and $\mathbb{R}^n$, respectively. Counting generator
weights only, the parameter count for each parameterization is
\begin{equation}
\begin{aligned}
P_{\mathrm{mHC}}
&= d(n^3+2n^2),\\
P_{\mathrm{CP}}
&= 3dr+7nr,\\
P_{\mathrm{Tucker}}
&= 7nr_n+3dr_d+2r_n^2r_d+r_n^3r_d.
\end{aligned}
\label{eq:temper_parameter_counts}
\end{equation}
Here, the rank $r$ is shared across the three TEMPER-CP generators, while TEMPER-Tucker uses
input- and output-stream rank, $r_n$, and feature rank, $r_d$. Thus, for fixed $d$ and
tensor ranks, dense mHC scales as $\mathcal{O}(n^3)$ in the number of streams,
whereas both TEMPER parameterizations scale as $\mathcal{O}(n)$. Figure~
\ref{fig:hyperconn_param_efficiency} reports the complete learnable parameter counts in hyper-connections, including biases, scalar gates, and RMSNorm scales.

\begin{table*}[!ht]
\centering
\small
\setlength{\tabcolsep}{2.1pt}
\caption{Additional parameters, training loss, validation BPB, and CORE score for hyper-connections variants with $D=12$ blocks and two residual connections per block. The $n=4$ and $n=8$ settings are respectively trained for 6{,}000 and 7{,}000 steps.}
\label{tab:loss}
\begin{tabular}{lcccccccc}
\toprule
Method 
& \multicolumn{4}{c}{$n=4$ (6000 steps)} 
& \multicolumn{4}{c}{$n=8$ (7000 steps)} \\
\cmidrule(lr){2-5} \cmidrule(lr){6-9}
& $\Delta$ Params (K) $\downarrow$
& Train Loss $\downarrow$ 
& Val BPB $\downarrow$ 
& CORE Score$\uparrow$
& $\Delta$ Params (K) $\downarrow$
& Train Loss $\downarrow$ 
& Val BPB $\downarrow$ 
& CORE Score$\uparrow$ \\
\midrule
Residual
& 0 & 2.705 & 0.816 & 0.182
& 0 & 2.645 & \underline{0.811} & 0.184 \\
mHC
& 1,844 & 2.714 & 0.819 & 0.177
& 11,946 & 2.627 & \underline{0.811} & 0.195\\
mHC-lite
& 2,434 & 2.724 & \underline{0.812} & 0.183
& 5,948,900 & -- & -- & -- \\
KromHC 
& 959 & \textbf{2.691} & \underline{0.812} & 0.186
& 3,392 & \textbf{2.612} & \textbf{0.807} & 0.195 \\
\rowcolor{blue!10}
TEMPER-CP \textbf{(Ours)}
& \textbf{186} & \underline{2.692} & \underline{0.812} & \underline{0.189}
& \textbf{376} & \underline{2.613} & \textbf{0.807} & \underline{0.201} \\
\rowcolor{blue!10}
TEMPER-Tucker \textbf{(Ours)}
& \underline{744} & 2.720 & \textbf{0.811} & \textbf{0.192}
& \underline{1,934} & \underline{2.613} & \textbf{0.807} & \textbf{0.206} \\
\bottomrule
\end{tabular}
\end{table*}

\begin{remark}
Prior work improves parameter efficiency mainly by reparameterizing
$\mathbf{A}^{\mathrm{res}}$. To our knowledge, TEMPER is the first to jointly
reparameterize the pre-, residual-, and post-routing generators, thus achieving
greater parameter efficiency while matching or improving performance.
\end{remark}

\subsection{Inductive Bias and Approximation Guarantees}
\label{sec:theoretical_properties}

As TEMPER-Tucker routes through the three compressed representations in
Eq.~\eqref{eq:tucker_compression}, its ranks restrict which
stream-feature directions affect routing and control how closely the
tensorized generators approximate their dense counterparts. We formalize
both properties below. TEMPER-CP is a special case of TEMPER-Tucker with a superdiagonal core.

\begin{theorem}\label{prop:tucker_degrees_of_freedom}
(\textbf{Rank and Null-space Dimension of the Tucker Compression}) For any of the three generators in Eq.~\eqref{eq:tucker-generator}, assume that $\mathbf{U} \in \mathbb{R}^{n
\times r_n}$ and $\mathbf{C} \in \mathbb{R}^{d \times r_d}$ have full
column rank. Given the normalized input $\hat{\mathbf{H}}\in\mathbb{R}^{n\times d}$, define
\begin{equation}
    L:\mathbb{R}^{n\times d}\to \mathbb{R}^{r_n\times r_d}, \quad
    L(\hat{\mathbf{H}})=\mathbf{U}^\top\hat{\mathbf{H}}\mathbf{C}.
\end{equation}
Then, $\operatorname{rank}(L)=r_n r_d$ and
$\dim \mathcal N(L)=nd-r_n r_d$.
\end{theorem}

Theorem~\ref{prop:tucker_degrees_of_freedom} shows that routing depends
only on an $r_n r_d$-dimensional learned subspace and is invariant to the
remaining directions. Increasing ranks enlarges this subspace, whereas low
ranks impose a stronger bottleneck that may suppress noisy or redundant
directions. At full multilinear rank, TEMPER-Tucker recovers the dense mHC
generator, so Tucker ranks interpolate between compact and dense
routing.

\begin{remark}
    Full-rank recovery is an expressivity guarantee rather than our design objective. When routing-relevant interactions lie in a low-dimensional structured subspace, compact ranks can provide sufficient routing capacity.
\end{remark}

\begin{lemma}\label{prop:dense_recovery_low_rank_approx}
\textbf{(Generator-to-logit Error Bound).}
Let $\mathcal{W}$ and $\widetilde{\mathcal{W}}$ denote any corresponding
dense and tensorized routing generators. For any normalized stream state
$\hat{\mathbf{H}}$, define
$\mathcal{Z}_{\mathcal{W}}(\hat{\mathbf{H}})
=\mathcal{W}\times_{1,2}\hat{\mathbf{H}}$, and similarly for
$\widetilde{\mathcal{W}}$. Then, the corresponding routing logits satisfy
\begin{equation}
\left\|
\mathcal{Z}_{\mathcal{W}}(\hat{\mathbf{H}})
-
\mathcal{Z}_{\widetilde{\mathcal{W}}}(\hat{\mathbf{H}})
\right\|_F
\le
\|\hat{\mathbf{H}}\|_F
\left\|
\mathcal{W}
-
\widetilde{\mathcal{W}}
\right\|_F .
\label{eq:approximation_bound}
\end{equation}
\end{lemma}

\begin{theorem}(\textbf{End-to-End Routed-Block Approximation Bound})
\label{prop:block_approx}
Let $\hat{\mathbf{H}}=\mathrm{RMSNorm}(\mathbf{H})$
be the normalized stream state, and $\mathbf{H}'$ and $\widetilde{\mathbf{H}}'$
the outputs of the dense and tensorized routed blocks. Let $f$ be the shared residual
transformation. Suppose that $f$ is $L_f$-Lipschitz and satisfies
$\|f(\mathbf{x})\|_2 \leq M_f$ for all relevant inputs, and that
$\Phi(\mathbf{Z})=\operatorname{SK}\!\left(\exp(\mathbf{Z})\right)$ is
$L_{\mathrm{SK}}$-Lipschitz on the residual-logit domain. Then,
\begin{equation}
\begin{aligned}
\left\|
\mathbf{H}'-\widetilde{\mathbf{H}}'
\right\|_F & \leq
|\alpha^{\mathrm{res}}| L_{\mathrm{SK}}
\|\hat{\mathbf{H}}\|_F
\|\mathbf{H}\|_F
\left\|\mathcal{W}^{\mathrm{res}}
-\widetilde{\mathcal{W}}^{\mathrm{res}}\right\|_F \\
&+
\frac{|\alpha^{\mathrm{post}}|}{2}
M_f
\|\hat{\mathbf{H}}\|_F
\left\|\mathcal{W}^{\mathrm{post}}
-\widetilde{\mathcal{W}}^{\mathrm{post}}\right\|_F
\\
&+
\frac{|\alpha^{\mathrm{pre}}|}{2}
\sqrt{n}\,L_f
\|\hat{\mathbf{H}}\|_F
\|\mathbf{H}\|_F
\left\|\mathcal{W}^{\mathrm{pre}}
-\widetilde{\mathcal{W}}^{\mathrm{pre}}\right\|_F .
\end{aligned}
\end{equation}
\end{theorem}

Lemma~\ref{prop:dense_recovery_low_rank_approx} quantifies how tensor
approximation error propagates to the routing logits. It shows that the discrepancy between dense and tensorized routing logits is bounded by the generator error $\left\|
\mathcal{W}
-
\widetilde{\mathcal{W}}
\right\|_F$ scaled by the normalized input magnitude $\|\hat{\mathbf{H}}\|_F$. Since RMS normalization
keeps this magnitude controlled, the resulting
logit discrepancy is governed primarily by the quality of tensor approximation.

Theorem~\ref{prop:block_approx} extends this result to the complete
routed block. Under the regularity assumptions, the block-output
difference is bounded by the approximation errors of the pre-, residual-,
and post-routing generators. Together, these results show how approximation errors are controlled as the routing parameterization moves from compact low-rank structure towards the dense unrestricted model. These guarantees
consider expressivity and approximation, rather than generalization, and 
Section~\ref{ablation} empirically examines why increasing the rank might not necessarily
improve performance. The proofs of theorems and lemma are given in the Supplement Section~1.

\section{Experiments}\label{sec:experiments}
\noindent
To evaluate the performance of TEMPER, we used the nanochat \cite{nanochat} backbone and apply different hyper-connections modules in different methods. The base model had $D=12$ Transformer blocks, hidden dimension $d=768$, $6$ attention heads, context length $2048$. We replaced residual branches in both attention and feed-forward network in every block, yielding $24$ residual modules per model. The pretraining used ClimbMix-400B dataset \cite{diao2025climb} with a $32{,}768$-token BPE tokenizer. $4\times$ NVIDIA A100 (80GB) GPUs are used for the experiments. All hyperparameters are provided in Supplement Section~4.

\subsection{Initialization}
We followed KromHC \cite{zhou2026kromhc} for routing-bias and scalar-gate
initialization, yielding stream-biased pre- and post-routing and near-identity residual mixing. Beyond this, all TEMPER-CP
factor entries were initialized independently from
$\mathcal{N}(0,r^{-1/2})$, where $r$ is the CP rank. For TEMPER-Tucker, all
factor and core entries are initialized independently from
$\mathcal{N}(0,r_{\max}^{-1/2})$, where $r_{\max}$ is the maximum mode rank. This rank-scaled initialization keeps factor magnitudes controlled and improves optimization stability.

\begin{table*}[!ht]
\centering
\small
\caption{Commonsense reasoning results for two stream counts. Best results are bolded. Second best results are underlined.}
\setlength{\tabcolsep}{4.5pt}
\begin{tabular}{clcccccccccccc}
\toprule
$n$ & Method
& Avg & HS & HS-ZS & PIQA & ARC-E & ARC-C & COPA & CSQA & OBQA & WG & WGrande & BoolQ \\
\midrule
& Residual (6000 steps)
& 49.0 & \underline{41.8} & 39.4 & \textbf{71.0} & \textbf{61.8} & \underline{33.0} & \textbf{65.0} & 25.6 & \textbf{35.0} & \underline{61.9} & 49.2 & 55.4 \\
&mHC
& 48.3 & 39.4 & 38.0 & 70.4 & 60.4 & 30.8 & 59.0 & 28.8 & 34.2 & 60.1 & 52.8 & 57.2 \\
&mHC-lite
& 48.9 & \textbf{42.6} & \underline{40.6} & \textbf{71.0} & 60.2 & \textbf{33.2} & 60.0 & 22.4 & 33.0 & 61.5 & \textbf{56.0} & 57.4 \\
&KromHC
& \underline{49.1} & 41.0 & \underline{40.6} & \underline{70.8} & \underline{60.8} & 32.8 & 60.0 & \underline{29.8} & 33.2 & 58.2 & \underline{53.0} & \textbf{60.0} \\
\rowcolor{blue!10}
&TEMPER-CP \textbf{(Ours)}
& 48.9 & 39.2 & 39.2 & 70.6 & 59.8 & 31.6 & 61.0 & \underline{34.4} & 32.4 & \textbf{62.3} & 52.4 & 55.2 \\
\rowcolor{blue!10}
\multirow{-5}{*}{4}
&TEMPER-Tucker \textbf{(Ours)}
& \textbf{50.0} & 41.0 & \textbf{41.6} & 68.2 & 60.2 & 30.8 & \underline{62.0} & \textbf{39.6} & \underline{34.8} & 60.1 & 52.2 & \underline{59.8} \\
\midrule
& Residual (7000 steps)
& 48.6 & 40.0 & \textbf{41.4} & 71.6 & \underline{63.2} & \underline{32.8} & \textbf{66.0} & 22.8 & 33.8 & 60.4 & 51.8 & 51.0 \\
&mHC
& 49.5 & 40.6 & \underline{41.2} & 72.2 & 61.6 & 32.0 & 64.0 & \underline{31.6} & 33.8 & \underline{61.9} & 50.4 & 55.4 \\
&mHC-lite
& -- & -- & -- & -- & -- & -- & -- & -- & -- & -- & -- & -- \\
&KromHC
& 49.8 & 40.6 & 40.6 & 72.4 & 62.6 & \textbf{33.2} & \underline{65.0} & 29.0 & 33.8 & \textbf{63.0} & 53.0 & 54.2 \\
\rowcolor{blue!10}
&TEMPER-CP \textbf{(Ours)}
& \underline{50.1} & \underline{41.6} & 41.0 & \textbf{73.0} & \textbf{63.4} & 32.2 & 63.0 & 29.8 & \underline{34.2} & 60.8 & \textbf{54.4} & \underline{57.6} \\
\rowcolor{blue!10}
\multirow{-5}{*}{8}
&TEMPER-Tucker \textbf{(Ours)}
& \textbf{51.3} & \textbf{42.0} & 40.8 & \underline{72.6} & 62.8 & 32.6 & \underline{65.0} & \textbf{42.2} & \textbf{35.6} & 58.2 & \underline{54.2} & \textbf{58.0} \\
\bottomrule
\end{tabular}
\label{tab:commonsense}
\end{table*}

\begin{table*}[t]
\centering
\caption{Language modeling results for the two stream counts. Best results are bolded. Second best results are underlined.}
\small
\setlength{\tabcolsep}{5pt}
\begin{tabular}{clccccccccccc}
\toprule
$n$ & Method
& Avg & Lamb & SQuAD & CoQA
& BBH-QA & BBH-CS & BBH-Op & BBH-Dyck & LSAT & LangID \\
\midrule
&Residual (6000 steps)
& 25.2 & \underline{35.2} & 25.2 & 19.4
& 36.4 & \underline{45.0} & 11.0 & 5.0 & 21.7 & 28.2 \\
&mHC
& 25.0 & 33.6 & 22.8 & 20.6
& 37.0 & 40.2 & \textbf{15.2} & 5.2 & 22.6 & 28.0 \\
&mHC-lite
& 25.3 & 33.2 & 25.8 & \underline{22.0}
& 34.6 & \underline{45.0} & 12.4 & 5.0 & 22.2 & 27.2 \\
&KromHC
& 25.7 & \textbf{35.4} & 25.0 & \textbf{22.4}
& 36.0 & 40.4 & \textbf{15.2} & 2.8 & \underline{25.2} & \underline{29.0} \\
\rowcolor{blue!10}
&TEMPER-CP \textbf{(Ours)}
& \textbf{27.1} & 34.8 & \textbf{27.2} & \textbf{22.4}
& \textbf{38.8} & \textbf{45.4} & \underline{13.8} & \underline{7.6} & \textbf{26.5} & 27.2 \\
\rowcolor{blue!10}
\multirow{-5}{*}{4}
&TEMPER-Tucker \textbf{(Ours)}
& \underline{25.9} & 33.4 & \underline{26.0} & 21.8 & \underline{37.4} & 44.4 & 10.5 & \textbf{8.6} & 21.7 & \textbf{29.4} \\
\midrule
& Residual (7000 steps)
& 26.9 & 34.2 & 25.8 & \textbf{23.0} & \textbf{39.6} & \textbf{46.2} & 13.8 & \textbf{9.6} & 23.9 & 25.6 \\
&mHC
& \underline{27.1} & 32.8 & 30.6 & \underline{22.4}
& \underline{38.0} & \underline{45.6} & \underline{13.8} & \underline{9.2} & \textbf{27.4} & 24.4 \\
&mHC-lite
& -- & -- & -- & --
& -- & -- & -- & -- & -- & -- \\
&KromHC
& 26.7 & \textbf{35.2} & \underline{31.8} & 20.8
& 33.6 & 42.4 & \underline{13.8} & \underline{9.2} & \underline{26.1} & \underline{27.4} \\
\rowcolor{blue!10}
&TEMPER-CP \textbf{(Ours)}
& \textbf{27.3} & \underline{35.0} & \textbf{34.4} & 22.2
& 37.8 & 42.2 & 12.9 & 7.4 & 25.7 & \textbf{28.2} \\
\rowcolor{blue!10}
\multirow{-5}{*}{8}
&TEMPER-Tucker \textbf{(Ours)}
& 26.4 & 33.8 & 31.2 & 20.8
& 33.4 & 43.4 & \textbf{14.3} & 9.0 & 25.2 & 26.4 \\
\bottomrule
\end{tabular}
\label{tab:language_modeling}
\end{table*}

\subsection{Evaluation Metrics}
Following KromHC \cite{zhou2026kromhc}, to measure efficiency and language-modeling quality, we report additional learnable parameters, training cross-entropy (CE) loss, and validation bits-per-byte (BPB). Additionally, we report the CORE score \cite{NEURIPS2024_19e4ea30}, the mean-centered accuracy over 22 tasks (20 shown here), as a summary of downstream performance (See Supplement Section 3). For each task, we report task accuracy (\%). Language-modeling tasks were scored by exact continuation match, while multiple-choice tasks select the option with the lowest continuation loss.

\subsection{Downstream Performance}
Tables~\ref{tab:loss}, \ref{tab:commonsense}, and
\ref{tab:language_modeling} report downstream performance for $n=4$ and $n=8$ hyper-connection streams. Across both settings, TEMPER provides the strongest performance--parameter efficiency trade-off among the hyper-connection baselines, matching or improving the downstream performance with substantially fewer additional parameters. mHC-lite results at $n=8$ are unavailable because its $\mathcal{O}(n!)$ residual-routing parameter growth led to an out-of-memory (OOM) error.

As shown in Table~\ref{tab:loss}, TEMPER-Tucker achieves the highest CORE score at both $n=4$ and $n=8$, improving over KromHC from $0.186$ to $0.192$ and from $0.195$ to $0.206$, respectively. At $n=8$, it uses $1.93$M additional parameters, compared with $3.39$M for KromHC and $11.95$M for mHC.

TEMPER-Tucker also achieves  the highest average accuracy on the commonsense and reasoning benchmark in Table~\ref{tab:commonsense}, reaching $50.0\%$ and $51.3\%$, respectively. At $n=8$, it outperformed the strongest existing baseline by $1.5\%$. For the language-modeling and BBH-oriented suite in Table~\ref{tab:language_modeling}, TEMPER-CP achieves the highest average accuracy in both stream counts, with $27.1\%$ for $n=4$ and $27.3\%$ for $n=8$. 

These results suggest an expressivity--regularization trade-off. TEMPER-Tucker's learned core captures richer cross-mode interactions, which may benefit tasks requiring diverse routing patterns, such as commonsense reasoning. In comparison, TEMPER-CP imposes a stronger separable prior with
greater parameter sharing, which may improve generalization on
language-modeling and BBH-oriented tasks.

\subsection{Ablation Studies}\label{ablation}

\paragraph{Frozen Tucker Core}
\begin{figure}[t]
\centering
\includegraphics[width=\columnwidth]{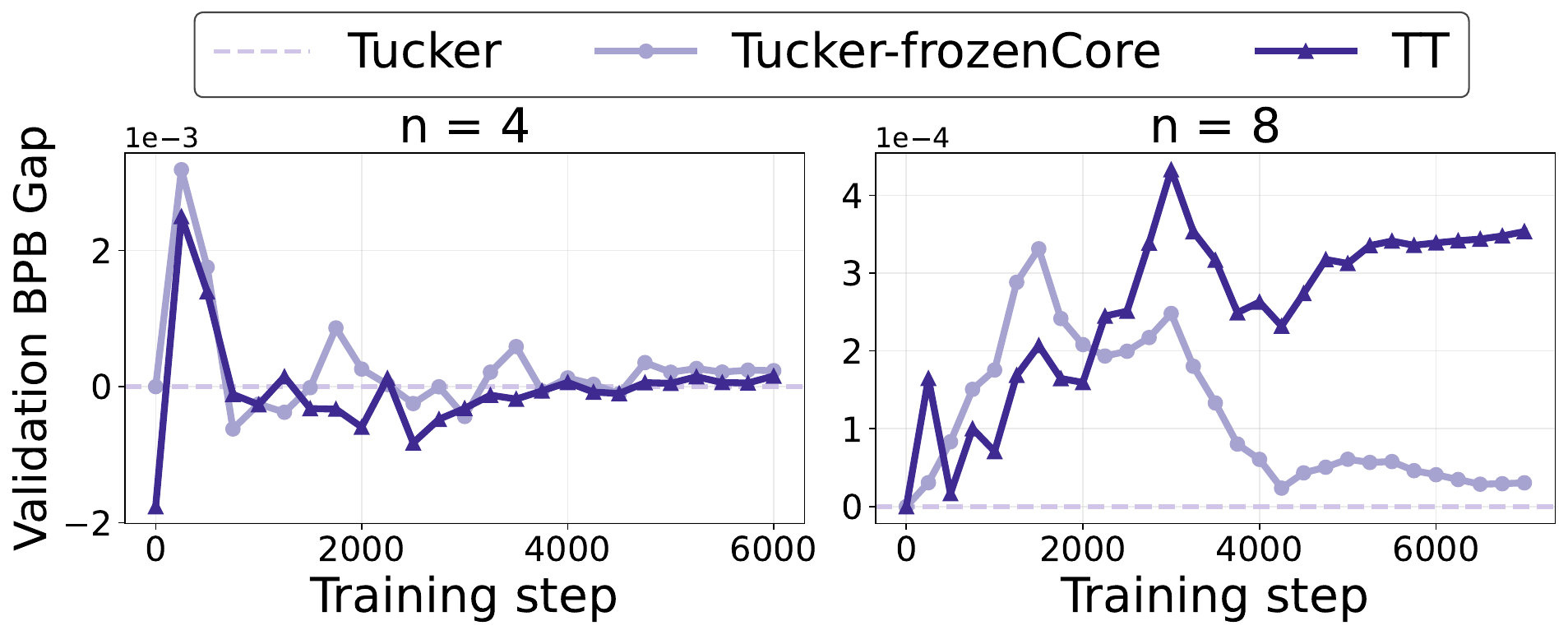}
\caption{Smoothed validation BPB gaps relative to TEMPER-Tucker for the frozen-core and TT variant at $n=4$ and $n=8$. Positive values indicate worse BPB.}
\label{fig:ablation_tucker_fc_tt_bpb}
\end{figure}
Eq.~\eqref{eq:tucker_logits} separates Tucker routing into two stages. First, factor matrices construct a compressed input-stream-feature and output-stream representation, and then the core tensor models interactions among the compressed coordinates. The frozen-core variant isolates these roles by keeping the factor matrices trainable but fixing the core at initialization. As shown in Figure~\ref{fig:ablation_tucker_fc_tt_bpb}, the frozen-core model has consistently worse validation BPB than the fully trainable Tucker model. Thus, learning the projection subspaces alone is insufficient. Adapting the interactions among those subspaces also contributes to the routing quality.

\paragraph{Effect of Tensor Network Topology}
Beyond the CP and Tucker decompositions, we evaluated a Tensor Train (TT) variant (See Supplement Section 2), given by
\begin{equation}
\mathcal{W}
=
\mathcal{G}_{1}\times_{3}^{1}\mathcal{G}_{2}\times_{4}^{1}\mathcal{G}_{3}\times_{5}^{1}
\cdots\times_{m+1}^{1}\mathcal{G}_{m},
\end{equation}
where $m=3$ for the pre- and post-branch generators and $m=4$ for the residual
generator. TT represents each generator as a chain of low-rank cores, rather
than Tucker's single global core. Figure~\ref{fig:ablation_tucker_fc_tt_bpb}
shows that TT has worse validation BPB than Tucker at both $n=4$ and $n=8$,
suggesting that a global core better suits this routing setting.

\paragraph{Tensorizing only $\mathbf{W}^{\text{res}}$}
\begin{figure}[t]
\centering
\includegraphics[width=\columnwidth]{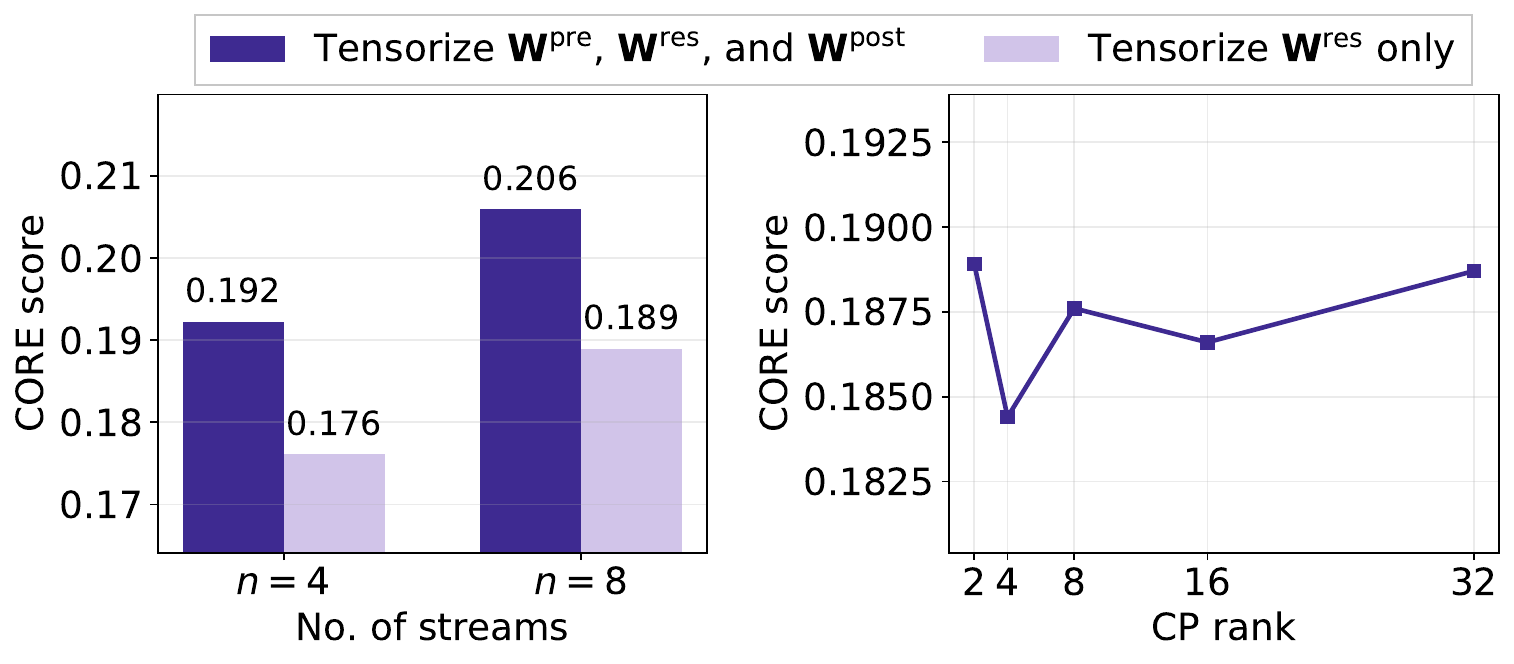}
\caption{Ablations on TEMPER routing design. \textbf{Left}: CORE score comparison at the final step for full TEMPER-Tucker against a residual-only variant that tensorizes $\mathbf{W}^{\mathrm{res}}$ only. \textbf{Right}: CORE score at final step for TEMPER-CP with ranks $r \in \{2,4,8,16,32\}$ at $n=4$.}
\label{fig:ablation_residual_cp_rank}
\end{figure}

Theorem~\ref{prop:block_approx} decomposes the block-output discrepancy into contributions from the three routing generators. To isolate the effect of the residual generator, we tensorized only $\mathbf{W}^{\text{res}}$, while keeping $\mathbf{W}^{\text{pre}}$ and $\mathbf{W}^{\text{post}}$ dense. As shown in Figure~\ref{fig:ablation_residual_cp_rank} (left), this variant underperformed full TEMPER-Tucker, whose CORE score increased from $0.176$ to $0.192$ at $n=4$, and from $0.189$ to $0.206$ at $n=8$. These results show that tensorizing the residual generator, $\mathbf{W}^{\text{res}}$, is beneficial, while jointly tensorizing all three generators yields the best performance.

\paragraph{Tensor ranks}
Theorem~\ref{prop:tucker_degrees_of_freedom} shows that tensor rank controls the dimension of the learned stream-feature subspace available for routing. Figure~\ref{fig:ablation_residual_cp_rank} (right) studies the effect of CP rank at $n=4$ under a fixed 6000-step budget. The final CORE score varied only from $0.184$ to $0.189$, with the best results at $r=2$. This weak performance sensitivity to rank suggests that the dense generator is over-parameterized for routing in this setting. A small number of shared low-rank components provides sufficient expressivity, while the resulting structural constraint may also act as an implicit regularizer.

\section{Conclusion}
We have introduced TEMPER, which jointly tensorizes pre-aggregation, residual-mixing, and post-redistribution routing generators in hyper-connections, reducing  parameter growth while retaining effective routing capacity. 
Our theoretical analysis characterizes routing capacity through tensor ranks and bounds how generator approximation errors propagate to routing logits and block outputs. 
Across language modeling and reasoning benchmarks, TEMPER has matched or outperformed prior hyper-connection methods, while requiring substantially fewer additional parameters, 
providing a scalable approach to expressive hyper-connections in LLMs.

\bibliography{aaai2027}
\end{document}